\documentclass[11pt,a4paper]{article}
\usepackage[hyperref]{acl2019}
\usepackage{times}
\usepackage{latexsym}

\usepackage{url}
\usepackage{graphicx}
\usepackage{wrapfig}
\usepackage{amsmath}
\usepackage{amssymb}
\usepackage{color,xcolor,colortbl}
\usepackage{enumitem}
\usepackage{algorithm}
\usepackage{algorithmic}
\usepackage[font={small}]{caption}
\usepackage{bm,bbm}
\usepackage{booktabs}
\usepackage{mathtools}
\usepackage{array}
\usepackage{subcaption}
\usepackage{pbox}
\usepackage{pifont}

\DeclareMathOperator*{\argmax}{arg\,max\,}
\newcommand\numberthis{\addtocounter{equation}{1}\tag{\theequation}}

\definecolor{darkblue}{rgb}{0.0, 0.0, 0.55}
\newenvironment{fontppl}{\fontfamily{ppl}\selectfont}{\par} % Palatino
\definecolor{input}{rgb}{0.63, 0.79, 0.95}
\definecolor{conv1}{rgb}{1.0, 0.89, 0.77}
\definecolor{maxpool}{rgb}{0.92, 0.3, 0.26}
\definecolor{dense}{rgb}{0.82, 0.62, 0.91}
\definecolor{relu}{rgb}{0.76, 0.33, 0.76}
\DeclareRobustCommand{\legendsquare}[1]{\textcolor{#1}{\rule{3ex}{1.5ex}}}
\usepackage{multirow}
\usepackage{comment}

\aclfinalcopy % Uncomment this line for the final submission
\def\aclpaperid{2143} %  Enter the acl Paper ID here

\title{Improving the Similarity Measure of Determinantal Point Processes for Extractive Multi-Document Summarization}

\author{Sangwoo Cho, Logan Lebanoff, Hassan Foroosh, Fei Liu\\
  Computer Science Department \\
  University of Central Florida, Orlando, FL 32816, USA \\
  {\texttt{\fontsize{10.5}{13}\selectfont \{swcho,loganlebanoff\}@knight.ucf.edu,  \{foroosh,feiliu\}@cs.ucf.edu}} \\}

\date{}

\begin{document}
\maketitle
\begin{abstract}

The most important obstacles facing multi-document summarization include excessive redundancy in source descriptions and the looming shortage of training data.
These obstacles prevent encoder-decoder models from being used directly, but optimization-based methods such as determinantal point processes (DPPs) are known to handle them well.
In this paper we seek to strengthen a DPP-based method for extractive multi-document summarization by presenting a novel similarity measure inspired by capsule networks. 
The approach measures redundancy between a pair of sentences based on surface form and semantic information. 
We show that our DPP system with improved similarity measure performs competitively, outperforming strong summarization baselines on benchmark datasets.
Our findings are particularly meaningful for summarizing documents created by multiple authors containing redundant yet lexically diverse expressions.\footnote{Our code and data are publicly available at {\fontsize{8}{9}\selectfont \url{https://github.com/ucfnlp/summarization-dpp-capsnet}}}

\end{abstract}

\section{Introduction}
\label{sec:intro}

Multi-document summarization is arguably one of the most important tools for information aggregation. 
It seeks to produce a succinct summary from a collection of textual documents created by multiple authors concerning a single topic~\cite{Nenkova:2011}.
The summarization technique has seen growing interest in a broad spectrum of domains that include summarizing product reviews~\cite{Gerani:2014,Yang:2018}, student survey responses~\cite{Luo:2015,Luo:2016:NAACL}, forum discussion threads~\cite{Ding:2015,Tarnpradab:2017}, and news articles about a particular event~\cite{Hong:2014}.
Despite the empirical success, most of the datasets remain small, and the cost of hiring human annotators to create ground-truth summaries for multi-document inputs can be prohibitive.

Impressive progress has been made on neural abstractive summarization using encoder-decoder models~\cite{Rush:2015,See:2017,Paulus:2017,Chen:2018:ACL}.
These models, nonetheless, are data-hungry and learn poorly from small datasets, as is often the case with multi-document summarization.
To date, studies have primarily focused on single-document summarization~\cite{See:2017,Celikyilmaz:2018,Kryscinski:2018} and sentence summarization~\cite{Nallapati:2016,Zhou:2017,Cao:2018,Song:2018} in part because parallel training data are abundant and they can be conveniently acquired from the Web.
Further, a notable issue with abstractive summarization is the reliability. 
These models are equipped with the capability of generating new words not present in the source.
With greater freedom of lexical choices, the system summaries can contain inaccurate factual details and falsified content that prevent them from staying ``true-to-original.''

In this paper we instead focus on an extractive method exploiting the determinantal point process (DPP; Kulesza and Taskar, 2012\nocite{Kulesza:2012}) for multi-document summarization.
DPP can be trained on small data, and because extractive summaries are free from manipulation, they largely remain true to the original. 
DPP selects a set of most representative sentences from the given source documents to form a summary, while maintaining high diversity among summary sentences.
It is one of a family of optimization-based summarization methods that performed strongest in previous summarization competitions~\cite{Gillick:2009:NAACL,Lin:2010:NAACL,Kulesza:2011}.

Diversity is an integral part of the DPP model. 
It is modelled by \emph{pairwise repulsion} between sentences. 
In this paper we exploit the capsule networks~\cite{Hinton:2018} to measure pairwise sentence (dis)similarity, then leverage DPP to obtain a set of diverse summary sentences.
Traditionally, the DPP method computes similarity scores based on the bag-of-words representation of sentences~\cite{Kulesza:2011} and with kernel methods~\cite{Gong:2014}.
These methods, however, are incapable of capturing lexical and syntactic variations in the sentences (e.g., paraphrases), which are ubiquitous in multi-document summarization data as the source documents are created by multiple authors with distinct writing styles.
We hypothesize that the recently proposed capsule networks, which learn high-level representations based on the orientational and spatial relationships of low-level components, can be a suitable supplement to model pairwise sentence similarity.

Importantly, we argue that predicting sentence similarity within the context of summarization has its uniqueness.
It estimates if two sentences contain redundant information based on both surface word form and their underlying semantics. 
E.g., the two sentences ``\emph{Snowstorm slams eastern US on Friday}'' and ``\emph{A strong wintry storm was dumping snow in eastern US after creating traffic havoc that claimed at least eight lives}'' are considered similar because they carry redundant information and cannot both be included in the summary.
These sentences are by no means semantically equivalent, nor do they exhibit a clear entailment relationship.
The task thus should be distinguished from similar tasks such as predicting natural language inference~\cite{Bowman:2015,Williams:2018}
or semantic textual similarity~\cite{Cer:2017}.
In this work, we describe a novel method to collect a large amount of sentence pairs that are deemed similar for summarization purpose.
We contrast this new dataset with those used for textual entailment for modeling sentence similarity and demonstrate its effectiveness on discriminating sentences and generating diverse summaries. 
The contributions of this work can be summarized as follows:

\begin{itemize}[topsep=3pt,itemsep=-1pt,leftmargin=*]

\item we present a novel method inspired by the determinantal point process for multi-document summarization.
The method includes a \emph{diversity} measure assessing the redundancy between sentences, and a \emph{quality} measure that indicates the importance of sentences. 
DPP extracts a set of summary sentences that are both representative of the document set and remain diverse;
 
\item we present the first study exploiting capsule networks for determining sentence similarity for summarization purpose. It is important to recognize that summarization places particular emphasis on measuring redundancy between sentences; and this notion of similarity is different from that of entailment and semantic textual similarity (STS);

\item our findings suggest that effectively modeling pairwise sentence similarity is crucial for increasing summary diversity and boosting summarization performance. Our DPP system with improved similarity measure performs competitively, outperforming strong summarization baselines on benchmark datasets.

\end{itemize}

\section{Related Work}
\label{sec:related}

Extractive summarization approaches are the most popular in real-world applications~\cite{Carbonell:1998,Daume:2006:ACL,Galanis:2010,Hong:2014,Yogatama:2015:EMNLP}.
These approaches focus on identifying representative sentences from a single document or set of documents to form a summary.
The summary sentences can  be optionally compressed to remove unimportant constituents such as prepositional phrases to yield a succinct summary~\cite{Knight:2002,Zajic:2007,Martins:2009,Kirkpatrick:2011,Thadani:2013,Wang:2013,Li:2013:EMNLP,Li:2014:EMNLP,Filippova:2015,Durrett:2016}.
Extractive summarization methods are mostly unsupervised or lightly-supervised using thousands of training examples. 
Given its practical importance, we explore an extractive method in this work for multi-document summarization.

It is not uncommon to cast summarization as a discrete optimization problem~\cite{Gillick:2009:NAACL,Takamura:2009,Lin:2010:NAACL,Hirao:2013}. 
In this formulation, a set of binary variables are used to indicate whether their corresponding source sentences are to be included in the summary.
The summary sentences are selected to maximize the coverage of important source content, while minimizing the summary redundancy and subject to a length constraint.
The optimization can be performed using an off-the-shelf tool such as Gurobi, IBM CPLEX, or via a greedy approximation algorithm.
Notable optimization frameworks include integer linear programming~\cite{Gillick:2009:NAACL}, determinantal point processes~\cite{Kulesza:2012}, submodular functions~\cite{Lin:2010:NAACL}, and minimum dominating set~\cite{Shen:2010}.
In this paper we employ the DPP framework because of its remarkable performance on various summarization problems~\cite{Zhang:2016:DPP}.

Recent years have also seen considerable interest in neural approaches to summarization.
In particular, neural extractive approaches focus on learning vector representations of source sentences; then based on these representations they determine if a source sentence is to be included in the summary~\cite{Cheng:2016,Yasunaga:2017,Nallapati:2017,Narayan:2018}.
Neural abstractive approaches usually include an encoder used to convert the entire source document to a continuous vector, and a decoder for generating an abstract word by word conditioned on the document vector~\cite{Paulus:2017,Tan:2017,Guo:2018:ACL,Kedzie:2018}.
These neural models, however, require large training data containing hundreds of thousands to millions of examples, which are still unavailable for the multi-document summarization task.
To date, most neural summarization studies are performed for single document summarization.

Extracting summary-worthy sentences from the source documents is important even if the ultimate goal is to generate abstracts.
Recent abstractive studies recognize the importance of separating ``salience estimation'' from ``text generation'' so as to reduce the amount of training data required by encoder-decoder models~\cite{Gehrmann:2018,Lebanoff:2018,Lebanoff:2019}.
An extractive method is often leveraged to identify salient source sentences, 
then a neural text generator rewrites the selected sentences into an abstract.
Our pursuit of the DPP method is especially meaningful in this context.
As described in the next section, DPP has an extraordinary ability to distinguish redundant descriptions, thereby avoiding passing redundant content to the abstractor that can cause an encoder-decoder model to fail.

\section{The DPP Framework}
\label{sec:dpp}

Let $\mathcal{Y} = \{1,2,\cdots,\textsf{N}\}$ be a ground set containing \textsf{N} items, corresponding to all sentences of the source documents.
Our goal is to identify a subset of items $Y \subseteq \mathcal{Y}$ that forms an extractive summary of the document set.
A determinantal point process (DPP; Kulesza and Taskar, 2012\nocite{Kulesza:2012}) defines a probability measure over all subsets of $\mathcal{Y}$ s.t.
\begin{align*}
\mathcal{P}(Y;L) &= \frac{\mbox{det}(L_Y)}{\mbox{det}(L + I)}, 
\numberthis\label{eq:p_y}\\
\sum_{Y \subseteq \mathcal{Y}}\mbox{det}(L_Y) &= \mbox{det}(L + I),
\numberthis\label{eq:sum_det_y}
\end{align*}
where 
$\mbox{det}(\cdot)$ is the determinant of a matrix;
$I$ is the identity matrix;
$L \in \mathbb{R}^{\textsf{N} \times \textsf{N}}$ is a positive semidefinite matrix, known as the $L$-ensemble; 
$L_{ij}$ measures the correlation between sentences $i$ and $j$;
and $L_Y$ is a submatrix of $L$ containing only entries indexed by elements of $Y$.
Finally, the probability of an extractive summary $Y \subseteq \mathcal{Y}$ is proportional to the determinant of the matrix $L_Y$ (Eq.~(\ref{eq:p_y})).
% I is a diagonal indicator matrix

\begin{figure}[t]
\centering
\includegraphics[width=0.8\linewidth]{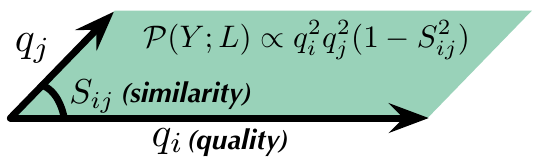}
\caption{The DPP model specifies the probability of a summary $\mathcal{P}(Y=\{i,j\};L)$ to be proportional to the squared volume of the space spanned by sentence vectors $i$ and $j$.}
\label{fig:dpp}
% \vspace{-0.1in}
\end{figure}

Kulesza and Taskar~\shortcite{Kulesza:2012} provide a decomposition of the $L$-ensemble matrix: 
$L_{ij} = q_i \cdot S_{ij} \cdot q_j$
where 
$q_i \in \mathbb{R}^+$ is a positive real number indicating the \emph{quality} of a sentence;
and $S_{ij}$ is a measure of \emph{similarity} between sentences $i$ and $j$.
This formulation separately models the sentence quality and pairwise similarity before combining them into a unified model.
Let $Y = \{i,j\}$ be a summary containing only two sentences $i$ and $j$, its probability $\mathcal{P}(Y;L)$ can be computed as
\begin{align*}
\mathcal{P}(Y=\{i,j\};L) &\propto \mbox{det}(L_Y)\\
&= 
\begin{vmatrix}
q_i S_{ii} q_i & q_i S_{ij} q_j\\
q_j S_{ji} q_i & q_j S_{jj} q_j
\end{vmatrix}
\\
&= q_i^2 \cdot q_j^2 \cdot (1-S_{ij}^2).
\numberthis\label{eq:p_y_toy}
\end{align*}
Eq.~(\ref{eq:p_y_toy}) indicates that, if sentence $i$ is of high quality, denoted by $q_i$, then any summary containing it will have high probability.
If two sentences $i$ and $j$ are similar to each other, denoted by $S_{ij}$, then any summary containing both sentences will have low probability. 
The summary $Y$ achieving the highest probability thus should contain a set of high-quality sentences while maintaining high diversity among the selected sentences (via pairwise repulsion).
$\mbox{det}(L_Y)$ also has a particular geometric interpretation as the squared volume of the space spanned by sentence vectors $i$ and $j$, where the quality measure indicates the length of the vector and the similarity indicates the angle between two vectors (Figure~\ref{fig:dpp}).

We adopt a feature-based approach to compute sentence quality: $q_i = \mbox{exp}(\boldsymbol{\theta}^\top \mathbf{x}_i)$.
In particular, $\mathbf{x}_i$ is a feature vector for sentence $i$ and $\boldsymbol{\theta}$ are the feature weights to be learned during training.
Kulesza and Taskar~\shortcite{Kulesza:2011} define sentence similarity as $S_{i,j} = \boldsymbol{\phi}_i^{\top} \boldsymbol{\phi}_j$, where $\left\lVert \boldsymbol{\phi}_i \right\rVert_2=1$ ($\forall i$) is a sentence TF-IDF vector. 
The model parameters $\boldsymbol{\theta}$ are optimized by maximizing the log-likelihood of training data (Eq.~(\ref{eq:mle})) and this objective can be optimized efficiently with subgradient
descent.\footnote{The sentence features include the length and position of a sentence, the cosine similarity between sentence and document TF-IDF vectors~\cite{Kulesza:2011}. We refrain from using sophisticated features to avoid model overfitting.}
{\medmuskip=1mu
\thinmuskip=1mu
\thickmuskip=1mu
\nulldelimiterspace=0pt
\scriptspace=0pt
\begin{align*}
\boldsymbol{\theta} = \argmax_{\boldsymbol{\theta}} \sum_{m=1}^M \log \mathcal{P}(\hat{Y}^{(m)}; L(\mathcal{Y}^{(m)}; \boldsymbol{\theta}))
\numberthis\label{eq:mle}
\end{align*}}

During training, we create the ground-truth extractive summary ($\hat{Y}$) for a document set based on human reference summaries (abstracts) using the following procedure.
At each iteration we select a source sentence sharing the longest common subsequence with the human reference summaries; the shared words are then removed from human summaries to avoid duplicates in future selection.
Similar methods are exploited by Nallapati et al.~\shortcite{Nallapati:2017} and Narayan et al.~\shortcite{Narayan:2018} to create ground-truth extractive summaries.
At test time, we perform inference using the learned DPP model to obtain a system summary ($Y$).
We implement a greedy method (Kulesza and Taskar, 2012\nocite{Kulesza:2012}) to iteratively add a sentence to the summary so that $\mathcal{P}(Y;L)$ yields the highest probability (Eq.~(\ref{eq:p_y})), until a summary length limit is reached.

For the DPP framework to be successful, the sentence similarity measure ($S_{ij}$) has to accurately capture if any two sentences contain redundant information.
This is especially important for multi-document summarization as redundancy is ubiquitous in source documents.
The source descriptions frequently contain redundant yet lexically diverse expressions such as sentential paraphrases where people write about the same event using distinct styles~\cite{Hu:2019}.
Without accurately modelling sentence similarity, redundant content can make their way into the summary and further prevent useful information from being included given the summary length limit. 
Existing cosine similarity measure between sentence TF-IDF vectors can be incompetent in modeling semantic relatedness.
In the following section we exploit the recently introduced capsule networks~\cite{Hinton:2018} to measure pairwise sentence similarity; it considers if two sentences share any words in common and more importantly the semantic closeness of sentence descriptions.

\begin{figure*}[t]
	\centering
	\includegraphics[width=0.95\linewidth]{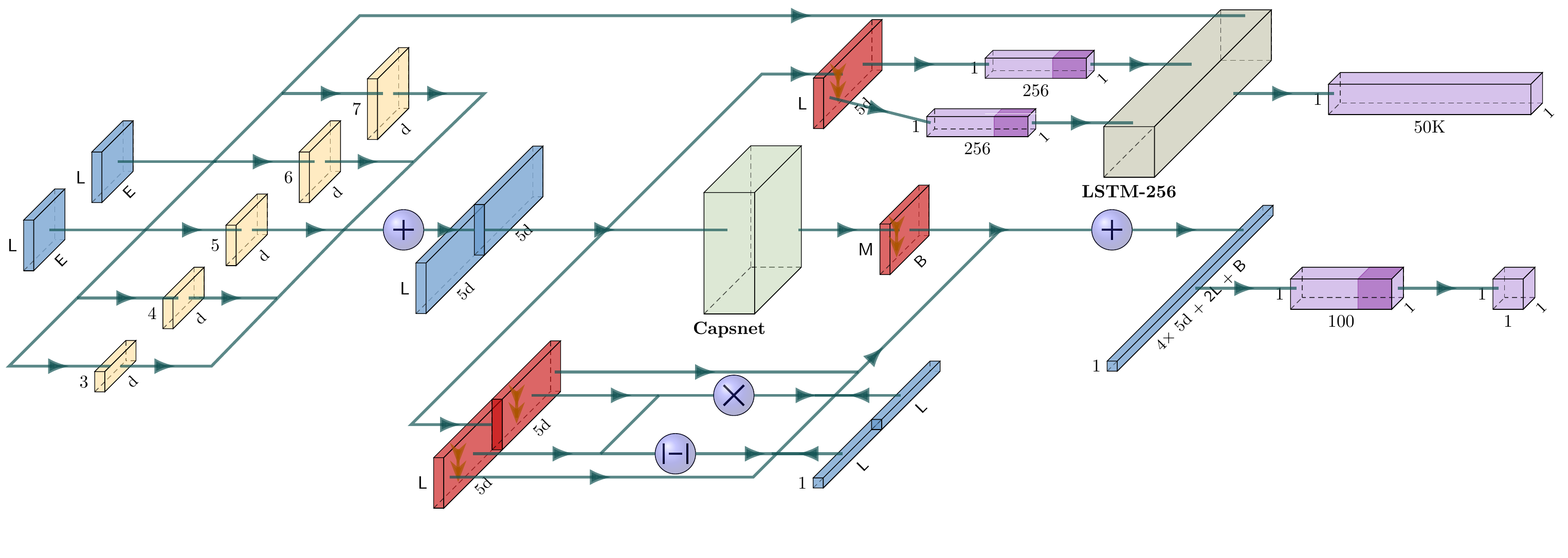}
    \vspace{-0.05in}
	\caption{
		The system architecture utilizing CapsNet for predicting sentence similarity.
		\legendsquare{input} denotes the inputs and intermediate outputs; \legendsquare{conv1} the convolutional layer; \legendsquare{maxpool} max-pooling layer; \legendsquare{dense} fully-connected layer; and \legendsquare{relu} ReLU activation.
	}
	\label{fig:architecture}
% 	\vspace{-0.1in}
\end{figure*}

\section{An Improved Similarity Measure}
\label{sec:capsnet}

Our goal is to develop an advanced similarity measure for pairs of sentences such that semantically similar sentences can receive high scores despite that they have very few words in common. 
E.g., ``\emph{Snowstorm slams eastern US on Friday}'' and ``\emph{A strong wintry storm was dumping snow in eastern US after creating traffic havoc that claimed at least eight lives}'' have only two words in common. Nonetheless, they contain redundant information and cannot both be included in the summary.

Let $\{\mathbf{x}^\textsf{a},\mathbf{x}^\textsf{b}\} \in \mathbb{R}^{\textsf{E} \times \textsf{L}}$ denote two sentences $\small\textsf{a}$ and $\small\textsf{b}$. 
Each consists of a sequence of word embeddings, where \textsf{E} is the embedding size and $\textsf{L}$ is the sentence length with zero-padding to the right for shorter sentences.
A convolutional layer with multiple filter sizes is first applied to each sentence to extract local features (Eq.~(\ref{eq:u_i_n})), 
where $\mathbf{x}_{i:i+k-1}^{\textsf{a}} \in \mathbb{R}^{k\textsf{E}}$ denotes a flattened embedding for position $i$ with a filter size $k$, and $\mathbf{u}_{i,k}^{\textsf{a}} \in \mathbb{R}^{d}$ is the resulting local feature for position $i$;
$f$ is a nonlinear activation function (e.g., ReLU);
$\{\mathbf{W}^u,\mathbf{b}^u\}$ are model parameters.
\begin{align*}
\mathbf{u}_{i,k}^{\textsf{a}} = f(\mathbf{W}^u\mathbf{x}_{i:i+k-1}^{\textsf{a}} + \mathbf{b}^u)
\numberthis\label{eq:u_i_n}
\end{align*}

We use $\mathbf{u}_{i}^{\textsf{a}} \in \mathbb{R}^{\textsf{D}}$ to denote the concatenation of local features generated using various filter sizes.
Following Kim et al.~\shortcite{Kim:2014}, we employ filter sizes $k \in \{3, 4, 5, 6, 7\}$ with an equal number of filters ($d$) for each size ($\textsf{D}=5d$).
After applying max-pooling to local features of all positions, we obtain a representation $\mathbf{u}^{\textsf{a}} = \mbox{max-pooling}(\mathbf{u}_{i}^{\textsf{a}}) \in \mathbb{R}^{\textsf{D}}$ for sentence $\small\textsf{a}$; and similarly we obtain $\mathbf{u}^{\textsf{b}} \in \mathbb{R}^{\textsf{D}}$ for sentence $\small\textsf{b}$.
It is not uncommon for state-of-the-art sentence similarity classifiers~\cite{Chen:2018:NLI} to concatenate the two sentence vectors, their absolute difference and element-wise product $[\mathbf{u}^{\textsf{a}}; \mathbf{u}^{\textsf{b}}; |\mathbf{u}^{\textsf{a}} - \mathbf{u}^{\textsf{b}}|; \mathbf{u}^{\textsf{a}} \circ \mathbf{u}^{\textsf{b}}]$, and feed this representation to a fully connected layer to predict if two sentences are similar.

Nevertheless, we conjecture that such representation may be insufficient to fully characterize the relationship between components of the sentences in order to model sentence similarity.
For example, the term ``\emph{snowstorm}'' in sentence $\small\textsf{a}$ is semantically related to ``\emph{wintry storm}'' and ``\emph{dumping snow}'' in sentence $\small\textsf{b}$; this low-level interaction indicates that the two sentences contain redundant information and it cannot be captured by the above model. 
Importantly, the capsule networks proposed by Hinton et al.~\shortcite{Hinton:2018} are designed to characterize the spatial and orientational relationships between low-level components.
We thus seek to exploit CapsNet to strengthen the capability of our system for identifying redundant sentences.

Let $\{\mathbf{u}_{i}^{\textsf{a}}, \mathbf{u}_{i}^{\textsf{b}}\}_{i=1}^{\textsf{L}} \in \mathbb{R}^\textsf{D}$ be low-level representations (i.e., capsules). 
We seek to transform them to high-level capsules $\{\mathbf{v}_{j}\}_{j=1}^{\textsf{M}} \in \mathbb{R}^\textsf{B}$ that characterize the interaction between low-level components.
Each low-level capsule $\mathbf{u}_{i} \in \mathbb{R}^\textsf{D}$ is multiplied by a linear transformation matrix to dedicate a portion of it, denoted by $\hat{\mathbf{u}}_{j|i} \in \mathbb{R}^\textsf{B}$, to the construction of a high-level capsule $j$ (Eq.~(\ref{eq:hat_u_j_i}));
where $\{\mathbf{W}_{ij}^v\} \in \mathbb{R}^{\textsf{D} \times \textsf{B}}$ are model parameters.
To reduce parameters and prevent overfitting, we further encourage sharing parameters over all low-level capsules, yielding $\mathbf{W}_{1j}^v = \mathbf{W}_{2j}^v = \cdots$, and the same parameter sharing is described in~\cite{Zhao:2018:Capsnet}.
By computing the weighted sum of $\hat{\mathbf{u}}_{j|i}$, whose weights $c_{ij}$ indicate the strength of interaction between a low-level capsule $i$ and a high-level capsule $j$,
we obtain an (unnormalized) capsule (Eq.~(\ref{eq:v_j})); 
we then apply a nonlinear squash function $g(\cdot)$ to normalize the length the vector to be less than 1, yielding $\mathbf{v}_j \in \mathbb{R}^{\textsf{B}}$.
\begin{align*}
\displaystyle
\hat{\mathbf{u}}_{j|i} &= \mathbf{W}_{ij}^v\mathbf{u}_i
\numberthis\label{eq:hat_u_j_i}\\
\mathbf{v}_j &= g\big(\sum_{i} c_{ij} \hat{\mathbf{u}}_{j|i}\big)
\numberthis\label{eq:v_j}
\end{align*}

Routing~\cite{Sabour:2017,Zhao:2019:Capsnet} aims to adjust the interaction weights ($c_{ij}$) using an iterative, EM-like method. 
Initially, we set $\{b_{ij}\}$ to be zero for all $i$ and $j$.
Per Eq.~(\ref{eq:c_i}), $\mathbf{c}_i$ becomes a uniform distribution indicating a low-level capsule $i$ contributes equally to all its upper level capsules.
After computing $\hat{\mathbf{u}}_{j|i}$ and $\mathbf{v}_j$ using Eq.~(\ref{eq:hat_u_j_i}-\ref{eq:v_j}), the weights $b_{ij}$ are updated according to the strength of interaction (Eq.~(\ref{eq:b_i_j})).
If $\hat{\mathbf{u}}_{j|i}$ agrees with a capsule $\mathbf{v}_j$, their interaction weight will be increased, and decreased otherwise.
This process is repeated for $r$ iterations to stabilize $c_{ij}$.
\begin{align*}
\mathbf{c}_i \leftarrow \mbox{softmax}(\mathbf{b}_i)
\numberthis\label{eq:c_i}\\
b_{ij} \leftarrow b_{ij} + \hat{\mathbf{u}}_{j|i}\mathbf{v}_j
\numberthis\label{eq:b_i_j}
\end{align*}

The high-level capsules $\{\mathbf{v}_j\}_{j=1}^{\textsf{M}}$ effectively encode spatial and orientational relationships of low-level capsules.
To identify the most prominent interactions, we apply max-pooling to all high-level capsules to produce $\mathbf{v} = \mbox{max-pooling}_j(\mathbf{v}_{j}) \in \mathbb{R}^{\textsf{B}}$. 
This representation $\mathbf{v}$, aimed to encode interactions between sentences $\small\textsf{a}$ and $\small\textsf{b}$, is concatenated with $[\mathbf{u}^{\textsf{a}}; \mathbf{u}^{\textsf{b}}; |\mathbf{u}^{\textsf{a}} - \mathbf{u}^{\textsf{b}}|; \mathbf{u}^{\textsf{a}} \circ \mathbf{u}^{\textsf{b}}]$ and binary vectors $[\mathbf{z}^{\textsf{a}};\mathbf{z}^{\textsf{b}}]$ that indicate if any word in sentence $\small\textsf{a}$ appears in sentence $\small\textsf{b}$ and vice versa;
they are used as input to a fully connected layer to predict if a pair of sentences contain redundant information.
Our loss function contains two components, including a binary cross-entropy loss indicating whether the prediction is correct or not, and a reconstruction loss for reconstructing a sentence $\small\textsf{a}$ conditioned on $\mathbf{u}^{\textsf{a}}$ by predicting one word at a time using a recurrent neural network, and similarly for sentence $\small\textsf{b}$.
A hyperparameter $\lambda$ is used to balance contributions from both sides. 
In Figure~\ref{fig:architecture} we present an overview of the system architecture, and hyper-parameters are described in the supplementary.

\section{Datasets}
\label{sec:data}

To our best knowledge, there is no dataset focusing on determining if two sentences contain redundant information. 
It is a nontrivial task in the context of multi-document summarization.
Further, we argue that the task should be distinguished from other semantic similarity tasks:
semantic textual similarity (STS; Cer et al., 2017\nocite{Cer:2017}) assesses to what degree two sentences are semantically equivalent to each other;
natural language inference (NLI; Bowman et al., 2015\nocite{Bowman:2015}) determines if one sentence (``hypothesis'') can be semantically inferred from the other sentence (``premise'').
Nonetheless, redundant sentences found in a set of source documents discussing a particular topic are not necessarily semantically equivalent or express an entailment relationship. 
We compare different datasets in \S\ref{sec:experiments}.

\vspace{0.06in}
\noindent\textbf{Sentence redundancy dataset}\quad
A novel dataset containing over 2 million sentence pairs is introduced in this paper for sentence redundancy prediction. 
We hypothesize that it is likely for a summary sentence and its most similar source sentence to contain redundant information. 
Because humans create summaries using generalization, paraphrasing, and other high-level text operations, a summary sentence and its source sentence can be semantically similar, yet contain diverse expressions.
Fortunately, such source/summary sentence pairs can be conveniently derived from single-document summarization data.
We analyze the CNN/Daily Mail dataset~\cite{Hermann:2015} that contains a massive collection of single news articles and their human-written summaries.
For each summary sentence, we identify its most similar source sentence by calculating the averaged R-1, R-2, and R-L F-scores~\cite{Lin:2004} between a source and summary sentences.
We consider a summary sentence to have no match if the score is lower than a threshold.
We obtain negative examples by randomly sampling two sentences from a news article. 
In total, our training / dev / test sets contain 2,084,798 / 105,936 / 86,144 sentence pairs and we make the dataset available to advance research on sentence redundancy. 

\vspace{0.08in}
\noindent\textbf{Summarization datasets}\quad
We evaluate our DPP-based system on benchmark multi-document summarization datasets.
The task is to create a succinct summary with up to 100 words from a cluster of 10 news articles discussing a single topic.
The DUC and TAC datasets~\cite{Over:2004,Dang:2008} have been used in previous summarization competitions.
In this paper we use DUC-03/04 and TAC-08/09/10/11 datasets that contain 60/50/48/44/46/44 document clusters respectively.
Four human reference summaries have been created for each document cluster by NIST assessors. 
Any system summaries are evaluated against human reference summaries using the ROUGE software~\cite{Lin:2004}\footnote{w/ options \textsf{-n 2 -m -w 1.2 -c 95 -r 1000 -l 100}}, where R-1, -2, and -SU4 respectively measure the overlap of unigrams, bigrams, unigrams and skip bigrams with a maximum distance of 4 words.
We report results on DUC-04 (trained on DUC-03) and TAC-11 (trained on TAC-08/09/10) that are often used as standard test sets~\cite{Hong:2014}.

\section{Experimental Results}
\label{sec:experiments}

In this section we discuss results that we obtained for multi-document summarization and determining redundancy between sentences.

\begin{table}[t]
\setlength{\tabcolsep}{5pt}
\renewcommand{\arraystretch}{1.2}
\centering
\begin{small}
\begin{tabular}{|l|rrr|}
\hline
& \multicolumn{3}{c|}{\textbf{DUC-04}}\\
\textbf{System} & \textbf{R-1} & \textbf{R-2} & \textbf{R-SU4} \\
\hline
\hline
Opinosis{\scriptsize~\cite{Ganesan:2010}} & 27.07 & 5.03 & 8.63\\
Extract+Rewrite{\scriptsize~\cite{Song:2018}} & 28.90 & 5.33 & 8.76 \\
Pointer-Gen{\scriptsize~\cite{See:2017}} & 31.43 & 6.03 & 10.01\\
SumBasic{\scriptsize~\cite{Vanderwende:2007}} & 29.48 & 4.25 & 8.64\\
KLSumm{\scriptsize~(Haghighi et al., 2009)\nocite{Haghighi:2009}} & 31.04 & 6.03 & 10.23 \\
LexRank{\scriptsize~\cite{Erkan:2004}} & 34.44 & 7.11 & 11.19 \\
Centroid{\scriptsize~\cite{Hong:2014}} & 35.49 & 7.80 & 12.02 \\
ICSISumm{\scriptsize~\cite{Gillick:2009:NAACL}} & 37.31 & 9.36 & 13.12 \\
\hline
DPP{\scriptsize~\cite{Kulesza:2011}}$\dagger$ & 38.10 & 9.14 & 13.40 \\
DPP-Capsnet (this work) & 38.25 & 9.22 & 13.40 \\
DPP-Combined (this work) & \textbf{39.35} & \textbf{10.14} & \textbf{14.15} \\
\hline
\end{tabular}
\end{small}
\caption{ROUGE results on the DUC-04 dataset. 
$\dagger$ indicates our reimplementation of Kulesza and Taskar~\shortcite{Kulesza:2011} system.
}
\label{tab:results_duc04}
\vspace{-0.1in}
\end{table}

\subsection{Summarization Results}
\label{sec:results_summarization}

We compare our system with a number of strong summarization baselines (Table~\ref{tab:results_duc04} and~\ref{tab:results_tac11}).
In particular, \textit{SumBasic}~\cite{Vanderwende:2007} is an extractive approach assuming words occurring frequently in a document cluster are more likely to be included in the summary;
\textit{KL-Sum}~\cite{Haghighi:2009} is a greedy approach adding a sentence to the summary to minimize KL divergence;
and \textit{LexRank}~\cite{Erkan:2004} is a graph-based approach computing sentence importance based on eigenvector centrality.

We additionally consider abstractive baselines to illustrate how well these systems perform on multi-document summarization: 
\textit{Opinosis}~\cite{Ganesan:2010} focuses on creating a word co-occurrence graph from the source documents and searching for salient graph paths to create an abstract;
\textit{Extract+Rewrite}~\cite{Song:2018} selects sentences using LexRank and condenses each sentence to a title-like summary;  
\textit{Pointer-Gen}~\cite{See:2017} seeks to generate abstracts by copying words from the source documents and generating novel words not present in the source text.

\begin{table}[t]
\setlength{\tabcolsep}{5pt}
\renewcommand{\arraystretch}{1.2}
\centering
\begin{small}
\begin{tabular}{|l|rrr|}
\hline
& \multicolumn{3}{c|}{\textbf{TAC-11}}\\
\textbf{System} & \textbf{R-1} & \textbf{R-2} & \textbf{R-SU4} \\
\hline
\hline
Opinosis{\scriptsize~\cite{Ganesan:2010}} & 25.15 & 5.12 & 8.12\\
Extract+Rewrite{\scriptsize~\cite{Song:2018}} & 29.07 & 6.11 & 9.20\\
Pointer-Gen{\scriptsize~\cite{See:2017}} & 31.44 & 6.40 & 10.20\\
% PG-MMR{\scriptsize~\cite{Lebanoff:2018}} & 37.17 & 10.92 & 14.04\\
SumBasic{\scriptsize~\cite{Vanderwende:2007}} & 31.58 & 6.06 & 10.06\\
KLSumm{\scriptsize~(Haghighi et al., 2009)\nocite{Haghighi:2009}} & 31.23 & 7.07 & 10.56 \\
LexRank{\scriptsize~\cite{Erkan:2004}} & 33.10 & 7.50 & 11.13 \\
\hline
DPP{\scriptsize~\cite{Kulesza:2011}}$\dagger$ & 36.95 & 9.83 & 13.57 \\
DPP-Capsnet (this work) & 36.61 & 9.30 & 13.09 \\
DPP-Combined (this work) & \textbf{37.30} & \textbf{10.13} & \textbf{13.78} \\
\hline
\end{tabular}
\end{small}
\caption{ROUGE results on the TAC-11 dataset. 
}
\label{tab:results_tac11}
\vspace{-0.1in}
\end{table}

\begin{table*}[t]
\setlength{\tabcolsep}{5pt}
\renewcommand{\arraystretch}{1.1}
\begin{scriptsize}
\begin{fontppl}

\begin{minipage}[b]{0.5\hsize}\centering
\begin{tabular}[t]{|p{2.9in}|}
\hline
\textbf{LexRank Summary}\\[2mm]
\textbullet\, The official, Dr. Charles J. Ganley, director of the office of nonprescription drug products at the Food and Drug Administration, said in an interview that the agency was ``revisiting the risks and benefits of the use of these drugs in children'' and that ``we're particularly concerned about the use of these drugs in children less than 2 years of age.''\\[1.8mm]

\textbullet\, The Consumer Healthcare Products Association, an industry trade group that has consistently defended the safety of pediatric cough and cold medicines, recommended in its own 156-page safety review, also released Friday, that the FDA consider mandatory warning labels saying that they should not be used in children younger than two.\\[1.8mm]

\textbullet\, Major makers of over-the-counter infant cough and cold medicines announced Thursday that they were voluntarily withdrawing their products from the market for fear that they could be misused by parents.\\[1.8mm]

\hline
\hline
\textbf{Pointer-Gen Summary}\\[2mm]
\textbullet\, Dr. Charles Ganley, a top food and drug administration official, said the agency was ``revisiting the risks and benefits of the use of these drugs in children,'' the director of the FDA's office of nonprescription drug products. \\[1.9mm]

\textbullet\, The FDA will formally consider revising labeling at a meeting scheduled for Oct. 18-19. \\[1.9mm]

\textbullet\, The withdrawal comes two weeks after reviewing reports of side effects over the last four decades, a 1994 study found that more than a third of all 3-year-olds in the United States were estimated to have been given to a child. \\[1.9mm]
\hline
\end{tabular}
\end{minipage}
\hfill
\begin{minipage}[b]{0.5\hsize}\centering
\begin{tabular}[t]{|p{3in}|}
\hline
\textbf{DPP-Combined Summary}\\[2mm]
\textbullet\, Johnson \& Johnson on Thursday voluntarily recalled certain infant cough and cold products, citing "rare" instances of misuse leading to overdoses.\\[1.8mm]

\textbullet\, Federal drug regulators have started a broad review of the safety of popular cough and cold remedies meant for children, a top official said Thursday.\\[1.8mm]

\textbullet\, Safety experts for the Food and Drug Administration urged the agency on Friday to consider an outright ban on over-the-counter, multi-symptom cough and cold medicines for children under 6.\\[1.8mm]

\textbullet\, Major makers of over-the-counter infant cough and cold medicines announced Thursday that they were voluntarily withdrawing their products from the market for fear that they could be misused by parents.\\[1.8mm]
\hline
\hline
\textbf{Human Reference Summary}\\[2mm]
\textbullet\, On March 1, 2007, the Food/Drug Administration (FDA) started a broad safety review of children's cough/cold remedies. \\[1.7mm]

\textbullet\, They are particularly concerned about use of these drugs by infants. \\[1.7mm]

\textbullet\, By September 28th, the 356-page FDA review urged an outright ban on all such medicines for children under six. \\[1.7mm]

\textbullet\, Dr. Charles Ganley, a top FDA official said ``We have no data on these agents of what's a safe and effective dose in Children.'' The review also stated that between 1969 and 2006, 123 children died from taking decongestants and antihistimines. \\[1.7mm]

\textbullet\, On October 11th, all such infant products were pulled from the markets. \\[1.7mm]
\hline
\end{tabular}
\end{minipage}

\end{fontppl}
\end{scriptsize}
\caption{Example system summaries and the human reference summary.
LexRank extracts long and comprehensive sentences that yield high graph centrality. 
Pointer-Gen (abstractive) has difficulty in generating faithful summaries (see the last bullet ``\emph{all 3-year-olds ... have been given to a child}'').
DPP is able to select a balanced set of representative and diverse sentences.
}
\label{tab:example_summaries}
\vspace{-0.05in}
\end{table*}
% TAC-11/D1119-A.M.100.D.B

Our DPP-based framework belongs to a strand of optimization-based methods. 
In particular, \textit{ICSISumm} (Gillick et al., 2009)\nocite{Gillick:2009:NAACL} formulates extractive summarization as integer linear programming; it identifies a globally-optimal set of sentences covering the most important concepts of the source documents;
\textit{DPP}~\cite{Kulesza:2011} selects an optimal set of sentences that are representative of the source documents and with maximum diversity, as determined by the determinantal point process.
Gong et al.~\shortcite{Gong:2014} show that the DPP performs well on summarizing both text and video.

We experiment with several variants of the DPP model: \textit{DPP-Capsnet} computes the similarity between sentences ($S_{ij}$) using the CapsNet described in Sec. \S\ref{sec:capsnet} and trained using our newly-constructed sentence redundancy dataset, whereas the default DPP framework computes sentence similarity as the cosine similarity of sentence TF-IDF vectors. 
\textit{DPP-Combined} linearly combines the cosine similarity with the CapsNet output using an interpolation coefficient determined on the dev set\footnote{The Capsnet coefficient $\lambda_c$ is selected to be 0.2 and 0.1 respectively for the DUC-04 and TAC-11 dataset.}.

\begin{table}
\setlength{\tabcolsep}{4pt}
\renewcommand{\arraystretch}{1.2}
\centering
\footnotesize
\begin{small}
\begin{tabular}{|l|r|r|r|r|}
\hline
\textbf{Dataset} & \textbf{Train} & \textbf{Dev} & \textbf{Test} & \textbf{Accu.}\\
\hline
\hline
\parbox[c][0.3in][c]{0.9in}{
STS-Benchmark\\
{\scriptsize~\cite{Cer:2017}}}
& 5,749 & 1,500 & 1,379 & {64.7}\%\\
\hline
\parbox[c][0.3in][c]{0.9in}{
SNLI
\\{\scriptsize~\cite{Bowman:2015}}} 
& 366,603 & 6,607 & 6,605 & {93.0}\% \\ 
\hline
\parbox[c][0.3in][c]{0.9in}{
Src-Summ Pairs \\
(this work)}
& 2,084,798 & 105,936 & 86,144 & \textbf{94.8}\%\\
\hline
\end{tabular}
\end{small}
\caption{Several sentence similarity datasets, and the performance of CapsNet on them.
SNLI discriminates between entailment and contradiction; 
STS is pretrained using Src-Summ pairs and fine-tuned on train split due to its small size.
}
\label{tab:sent_pairs}
% \vspace{-0.15in}
\end{table}

Table~\ref{tab:results_duc04} and~\ref{tab:results_tac11} illustrate the summarization results we have obtained for the DUC-04 and TAC-11 datasets.
Our DPP methods perform superior to both extractive and abstractive baselines, indicating the effectiveness of optimization-based methods for extractive multi-document summarization.
The DPP optimizes for summary sentence selection to maximize their content coverage and diversity, expressed as the squared volume of the space spanned by the selected sentences.

\begin{table}[t]
\setlength{\tabcolsep}{5pt}
\renewcommand{\arraystretch}{1.2}
\centering
\begin{small}
\begin{tabular}{|p{2.9in}|}
\hline
\textbf{STS-Benchmark} \textbf{(a)} \emph{Four girls happily walk down a sidewalk.}
\textbf{(b)} \emph{Three young girls walk down a sidewalk.} \ding{55}\\
\hline
\hline
\textbf{SNLI} \textbf{(a)} \emph{3 young man in hoods standing in the middle of a quiet street facing the camera.} \textbf{(b)} \emph{Three hood wearing people pose for a picture.} \ding{51}\\
\hline
\hline
\textbf{Src-Summ Pairs} \textbf{(a)} \emph{He ended up killing five girls and wounding five others before killing himself.} \textbf{(b)} \emph{Nearly four months ago, a milk delivery-truck driver lined up 10 girls in a one-room schoolhouse in this Amish farming community and opened fire, killing five of them and wounding five others before turning the gun on himself.} \ding{51}\\
\hline
\end{tabular}
\end{small}
\caption{Example positive (\ding{51}) and negative (\ding{55}) sentence pairs from the semantic similarity datasets.
}
\label{tab:sent_pairs_example}
\vspace{-0.15in}
\end{table}

Further, we observe that the DPP system with combined similarity metrics yields the highest performance, achieving 10.14\% and 10.13\% F-scores respectively on DUC-04 and TAC-11.
This finding suggests that the cosine similarity of sentence TF-IDF vectors and the CapsNet semantic similarity successfully complement each other to provide the best overall estimate of sentence redundancy.
A close examination of the system outputs reveal that important topical words (e.g., ``\emph{\$3 million}'') that are frequently discussed in the document cluster can be crucial for determining sentence redundancy, because sentences sharing the same topical words are more likely to be considered redundant.
While neural models such as the CapsNet rarely explicitly model word frequencies, the TF-IDF sentence representation is highly effective in capturing topical terms.

In Table~\ref{tab:example_summaries} we show example system summaries and a human-written reference summary.
We observe that LexRank tends to extract long and comprehensive sentences that yield high graph centrality;
the abstractive pointer-generator networks, despite the promising results, can sometimes fail to generate meaningful summaries (e.g., ``\emph{a third of all 3-year-olds $\cdots$ have been given to a child}'').
In contrast, our DPP method is able to select a balanced set of representative and diverse summary sentences.
We next compare several semantic similarity datasets to gain a better understanding of modeling sentence redundancy for summarization.

\subsection{Sentence Similarity}
\label{sec:results_redundancy}

We compare three standard datasets used for semantic similarity tasks, including \emph{SNLI}~\cite{Bowman:2015}, used for natural language inference, \emph{STS-Benchmark}~\cite{Cer:2017} for semantic equivalence, and our newly-constructed \emph{Src-Summ} sentence pairs. 
Details are presented in Table~\ref{tab:sent_pairs}.

We observe that CapsNet achieves the highest prediction accuracy of 94.8\% on the \emph{Src-Summ} dataset and it yields similar performance on \emph{SNLI}, indicating the effectiveness of CapsNet on characterizing semantic similarity.
\emph{STS} appears to be a more challenging task, where CapsNet yields 64.7\% accuracy.
Note that we perform two-way classification on SNLI to discriminate entailment and contradiction. The STS dataset is too small to be used to train CapsNet without overfitting, we thus pre-train the model on \emph{Src-Summ} pairs, and use the train split of \emph{STS} to fine-tune parameters.

Table~\ref{tab:sent_pairs_example} shows example positive and negative sentence pairs from the \emph{STS}, \emph{SNLI}, and \emph{Src-Summ} datasets.
The \emph{STS} and \emph{SNLI} datasets are constructed by human annotators to test a system's capability of learning sentence representations.
The sentences can share very few words in common but still express an entailment relationship (positive); or the sentences can share a lot of words in common yet they are semantically distinct (negative).
These cases are usually not seen in summarization datasets containing clusters of documents discussing single topics.
The \emph{Src-Summ} dataset successfully strike a balance between sharing common words yet containing diverse expressions. It is thus a good fit for training classifiers to detect sentence redundancy.

\begin{figure}[t!]	
	\begin{minipage}[b]{.32\linewidth}
		\centering
		\centerline{\includegraphics[width=1in]{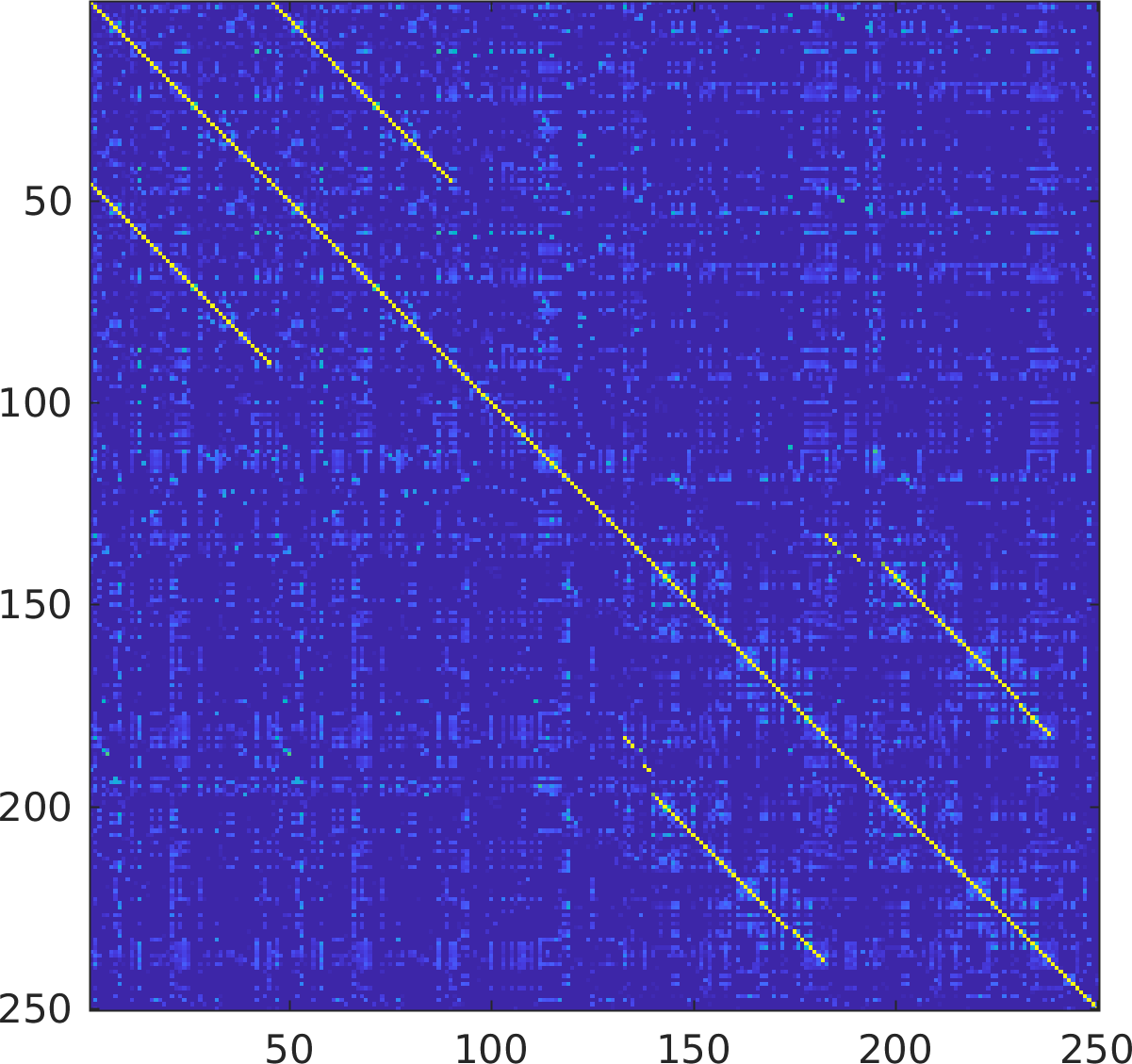}}
		\label{fig:sim_cosine}
	\end{minipage}
	%\hfill
	\begin{minipage}[b]{.32\linewidth}
		\centering
		\centerline{\includegraphics[width=1in]{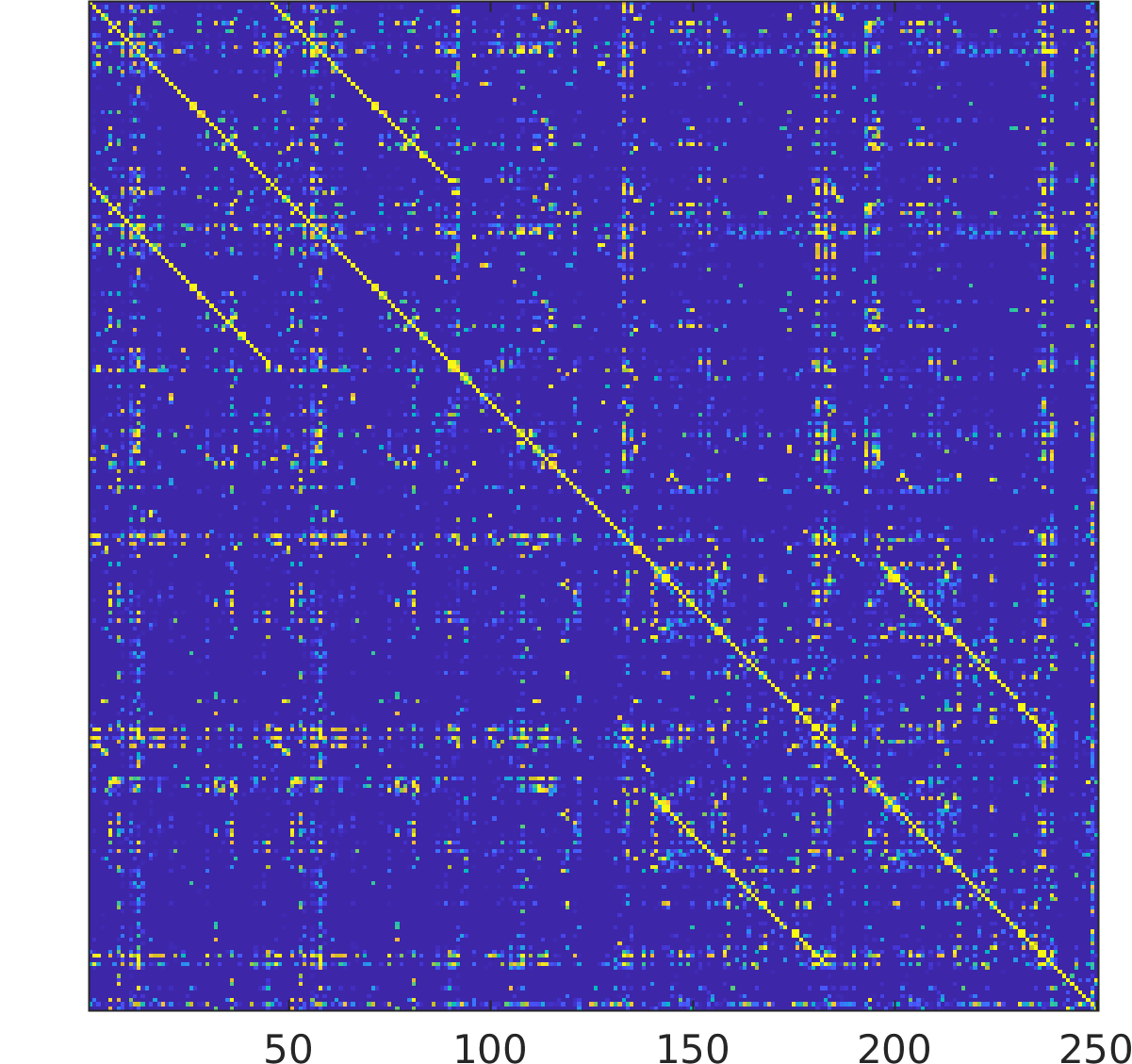}}
		\label{fig:sim_cnndm}
	\end{minipage}
	%\hfill
	\begin{minipage}[b]{.32\linewidth}
    	\centering
    	\centerline{\includegraphics[width=1in]{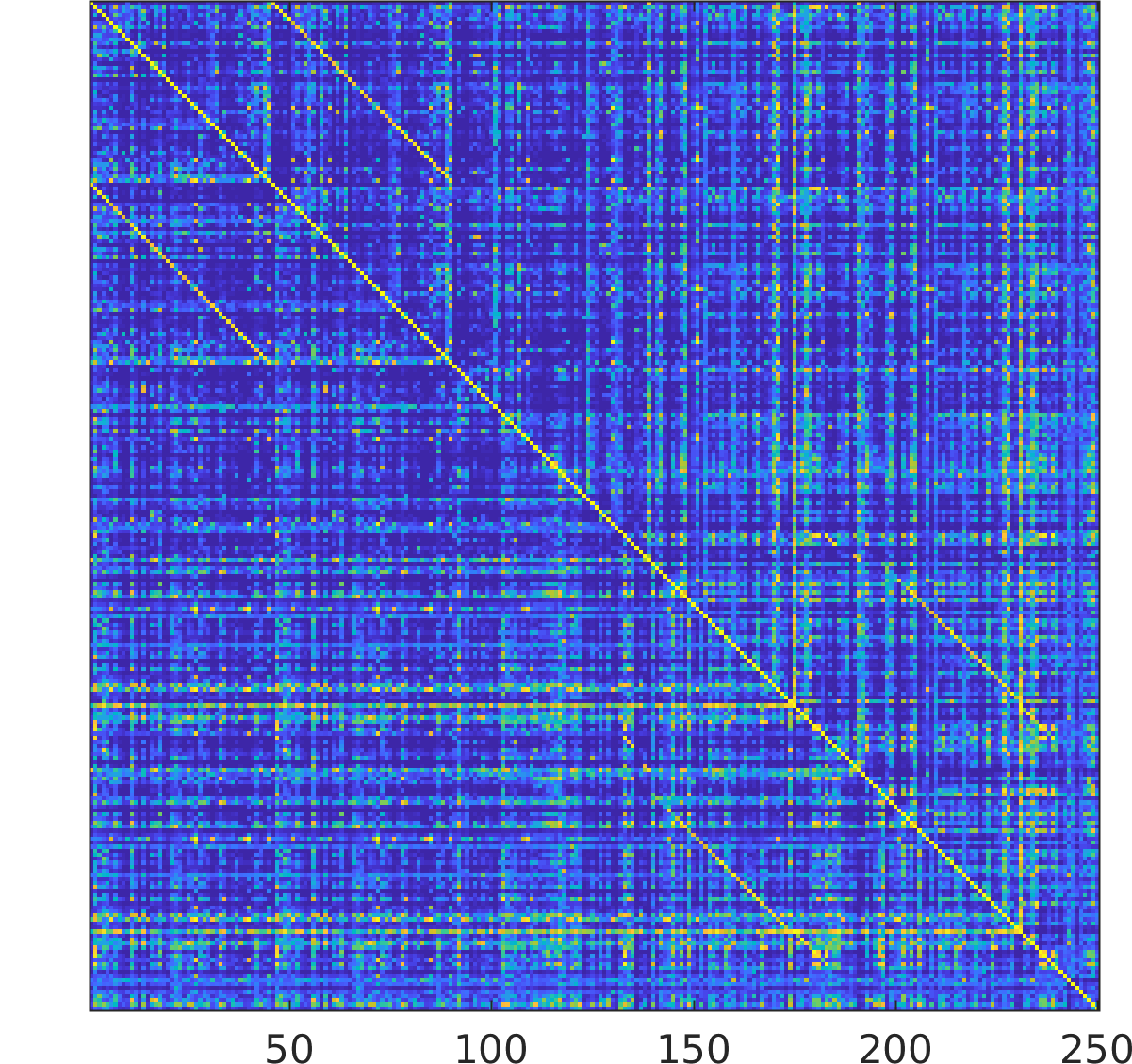}}
    	\label{fig:sim_snli}
	\end{minipage}
	\caption{
	Heatmaps for topic D31008 of DUC-04 (cropped to 200 sentences). It shows the cosine similarity of sentence TF-IDF vectors (\emph{Cosine}, left), and the CapsNet output trained respectively on \emph{SNLI} (right) and \emph{Src-Summ} (middle) datasets. The short off-diagonal lines are near-identical sentences found in the document cluster.
	}
	\label{fig:similarity_matrix}
\vspace{-0.1in}
\end{figure}

Figure~\ref{fig:similarity_matrix} compares heatmaps generated by computing cosine similarity of sentence TF-IDF vectors (\emph{Cosine}), and training CapsNet on \emph{SNLI} and \emph{Src-Summ} datasets respectively.
We find that the \emph{Cosine} similarity scores are relatively strict, as a vast majority of sentence pairs are assigned zero similarity, because these sentences have no word overlap.
At the other extreme, \emph{CapsNet+SNLI} labels a large quantity of sentence pairs as false positives, because its training data frequently contain sentences that share few words in common but nonetheless are positive, i.e., expressing an entailment relationship.
The similarity scores generated by \emph{CapsNet+SrcSumm} are more moderate comparing to \emph{CapsNet+SNLI} and \emph{Cosine}, suggesting the appropriateness of using \emph{Src-Summ} sentence pairs for estimating sentence redundancy.

\section{Conclusion}

We strengthen a DPP-based multi-document summarization system with improved similarity measure inspired by capsule networks for determining sentence redundancy.
We show that redundant sentences not only have common words but they can be semantically similar with little word overlap. 
Both aspects should be modelled in calculating pairwise sentence similarity.
Our system yields competitive results on benchmark datasets surpassing strong summarization baselines.

\section*{Acknowledgments}
The authors are grateful to the reviewers for their insightful feedback. We would also like to extend our thanks to Boqing Gong, Xiaodan Zhu and Fei Sha for useful discussions.

\bibliography{capsnet,summ,abs_summ,fei}

\begin{thebibliography}{69}
\expandafter\ifx\csname natexlab\endcsname\relax\def\natexlab#1{#1}\fi

\bibitem[{Berg-Kirkpatrick et~al.(2011)Berg-Kirkpatrick, Gillick, and
  Klein}]{Kirkpatrick:2011}
Taylor Berg-Kirkpatrick, Dan Gillick, and Dan Klein. 2011.
\newblock \href {https://dl.acm.org/citation.cfm?id=2002534} {Jointly learning
  to extract and compress}.
\newblock In \emph{Proceedings of the Annual Meeting of the Association for
  Computational Linguistics (ACL)}.

\bibitem[{Bowman et~al.(2015)Bowman, Angeli, Potts, and Manning}]{Bowman:2015}
Samuel~R. Bowman, Gabor Angeli, Christopher Potts, and Christopher~D. Manning.
  2015.
\newblock \href {https://nlp.stanford.edu/pubs/snli_paper.pdf} {A large
  annotated corpus for learning natural language inference}.
\newblock In \emph{Proceedings of the Conference on Empirical Methods in
  Natural Language Processing (EMNLP)}.

\bibitem[{Cao et~al.(2018)Cao, Wei, Li, and Li}]{Cao:2018}
Ziqiang Cao, Furu Wei, Wenjie Li, and Sujian Li. 2018.
\newblock \href {https://arxiv.org/abs/1711.04434} {Faithful to the original:
  {F}act aware neural abstractive summarization}.
\newblock In \emph{Proceedings of the AAAI Conference on Artificial
  Intelligence (AAAI)}.

\bibitem[{Carbonell and Goldstein(1998)}]{Carbonell:1998}
Jaime Carbonell and Jade Goldstein. 1998.
\newblock \href {https://dl.acm.org/citation.cfm?id=291025} {The use of {MMR},
  diversity-based reranking for reordering documents and producing summaries}.
\newblock In \emph{Proceedings of the International ACM SIGIR Conference on
  Research and Development in Information Retrieval (SIGIR)}.

\bibitem[{Celikyilmaz et~al.(2018)Celikyilmaz, Bosselut, He, and
  Choi}]{Celikyilmaz:2018}
Asli Celikyilmaz, Antoine Bosselut, Xiaodong He, and Yejin Choi. 2018.
\newblock \href {https://arxiv.org/pdf/1803.10357.pdf} {Deep communicating
  agents for abstractive summarization}.
\newblock In \emph{Proceedings of the North American Chapter of the Association
  for Computational Linguistics (NAACL)}.

\bibitem[{Cer et~al.(2017)Cer, Diab, Agirre, Lopez-Gazpio, , and
  Specia}]{Cer:2017}
Daniel Cer, Mona Diab, Eneko Agirre, Inigo Lopez-Gazpio, , and Lucia Specia.
  2017.
\newblock \href {https://www.aclweb.org/anthology/S17-2001} {{SemEval-2017}
  task 1: {S}emantic textual similarity multilingual and cross-lingual focused
  evaluation}.
\newblock In \emph{Proceedings of the 11th International Workshop on Semantic
  Evaluations (SemEval)}.

\bibitem[{Chen et~al.(2018)Chen, Zhu, Ling, Inkpen, and Wei}]{Chen:2018:NLI}
Qian Chen, Xiaodan Zhu, Zhen-Hua Ling, Diana Inkpen, and Si~Wei. 2018.
\newblock \href {https://aclweb.org/anthology/P18-1224} {Neural natural
  language inference models enhanced with external knowledge}.
\newblock In \emph{Proceedings of the Annual Meeting of the Association for
  Computational Linguistics (ACL)}.

\bibitem[{Chen and Bansal(2018)}]{Chen:2018:ACL}
Yen-Chun Chen and Mohit Bansal. 2018.
\newblock \href {https://arxiv.org/abs/1805.11080} {Fast abstractive
  summarization with reinforce-selected sentence rewriting}.
\newblock In \emph{Proceedings of the Annual Meeting of the Association for
  Computational Linguistics (ACL)}.

\bibitem[{Cheng and Lapata(2016)}]{Cheng:2016}
Jianpeng Cheng and Mirella Lapata. 2016.
\newblock \href {https://www.aclweb.org/anthology/P16-1046} {Neural
  summarization by extracting sentences and words}.
\newblock In \emph{Proceedings of ACL}.

\bibitem[{Dang and Owczarzak(2008)}]{Dang:2008}
Hoa~Trang Dang and Karolina Owczarzak. 2008.
\newblock \href
  {https://www.nist.gov/publications/overview-tac-2008-update-summarization-task}
  {Overview of the {TAC} 2008 update summarization task}.
\newblock In \emph{Proceedings of Text Analysis Conference (TAC)}.

\bibitem[{Daum{\'e}~III and Marcu(2006)}]{Daume:2006:ACL}
Hal Daum{\'e}~III and Daniel Marcu. 2006.
\newblock \href {https://arxiv.org/abs/0907.1814} {Bayesian query-focused
  summarization}.
\newblock In \emph{Proceedings of the 44th Annual Meeting of the Association
  for Computational Linguistics (ACL)}.

\bibitem[{Ding and Jiang(2015)}]{Ding:2015}
Ying Ding and Jing Jiang. 2015.
\newblock \href {https://www.aclweb.org/anthology/R15-1020} {Towards opinion
  summarization from online forums}.
\newblock In \emph{Proceedings of the International Conference Recent Advances
  in Natural Language Processing (RANLP)}.

\bibitem[{Durrett et~al.(2016)Durrett, Berg-Kirkpatrick, and
  Klein}]{Durrett:2016}
Greg Durrett, Taylor Berg-Kirkpatrick, and Dan Klein. 2016.
\newblock \href {https://arxiv.org/abs/1603.08887} {Learning-based
  single-document summarization with compression and anaphoricity constraints}.
\newblock In \emph{Proceedings of the Association for Computational Linguistics
  (ACL)}.

\bibitem[{Erkan and Radev(2004)}]{Erkan:2004}
G\"{u}nes Erkan and Dragomir~R. Radev. 2004.
\newblock \href {https://www.aaai.org/Papers/JAIR/Vol22/JAIR-2214.pdf}
  {{LexRank}: {G}raph-based lexical centrality as salience in text
  summarization}.
\newblock \emph{Journal of Artificial Intelligence Research}.

\bibitem[{Filippova et~al.(2015)Filippova, Alfonseca, Colmenares, Kaiser, and
  Vinyals}]{Filippova:2015}
Katja Filippova, Enrique Alfonseca, Carlos Colmenares, Lukasz Kaiser, and Oriol
  Vinyals. 2015.
\newblock \href
  {https://static.googleusercontent.com/media/research.google.com/en//pubs/archive/43852.pdf}
  {Sentence compression by deletion with lstms}.
\newblock In \emph{Proceedings of the Conference on Empirical Methods in
  Natural Language Processing (EMNLP)}.

\bibitem[{Galanis and Androutsopoulos(2010)}]{Galanis:2010}
Dimitrios Galanis and Ion Androutsopoulos. 2010.
\newblock \href {https://www.aclweb.org/anthology/N10-1131} {An extractive
  supervised two-stage method for sentence compression}.
\newblock In \emph{Proceedings of NAACL-HLT}.

\bibitem[{Ganesan et~al.(2010)Ganesan, Zhai, and Han}]{Ganesan:2010}
Kavita Ganesan, ChengXiang Zhai, and Jiawei Han. 2010.
\newblock \href {https://www.aclweb.org/anthology/C10-1039} {Opinosis: {A}
  graph-based approach to abstractive summarization of highly redundant
  opinions}.
\newblock In \emph{Proceedings of the International Conference on Computational
  Linguistics (COLING)}.

\bibitem[{Gehrmann et~al.(2018)Gehrmann, Deng, and Rush}]{Gehrmann:2018}
Sebastian Gehrmann, Yuntian Deng, and Alexander~M. Rush. 2018.
\newblock \href {https://arxiv.org/abs/1808.10792} {Bottom-up abstractive
  summarization}.
\newblock In \emph{Proceedings of the Conference on Empirical Methods in
  Natural Language Processing (EMNLP)}.

\bibitem[{Gerani et~al.(2014)Gerani, Mehdad, Carenini, Ng, and
  Nejat}]{Gerani:2014}
Shima Gerani, Yashar Mehdad, Giuseppe Carenini, Raymond~T. Ng, and Bita Nejat.
  2014.
\newblock \href {https://emnlp2014.org/papers/pdf/EMNLP2014168.pdf}
  {Abstractive summarization of product reviews using discourse structure}.
\newblock In \emph{Proceedings of the Conference on Empirical Methods in
  Natural Language Processing (EMNLP)}.

\bibitem[{Gillick and Favre(2009)}]{Gillick:2009:NAACL}
Dan Gillick and Benoit Favre. 2009.
\newblock \href {https://dl.acm.org/citation.cfm?id=1611640} {A scalable global
  model for summarization}.
\newblock In \emph{Proceedings of the NAACL Workshop on Integer Linear
  Programming for Natural Langauge Processing}.

\bibitem[{Gong et~al.(2014)Gong, Chao, Grauman, and Sha}]{Gong:2014}
Boqing Gong, Wei-Lun Chao, Kristen Grauman, and Fei Sha. 2014.
\newblock \href
  {https://papers.nips.cc/paper/5413-diverse-sequential-subset-selection-for-supervised-video-summarization.pdf}
  {Diverse sequential subset selection for supervised video summarization}.
\newblock In \emph{Proceedings of Neural Information Processing Systems
  (NIPS)}.

\bibitem[{Guo et~al.(2018)Guo, Pasunuru, and Bansal}]{Guo:2018:ACL}
Han Guo, Ramakanth Pasunuru, and Mohit Bansal. 2018.
\newblock \href {https://arxiv.org/abs/1805.11004} {Soft, layer-specific
  multi-task summarization with entailment and question generation}.
\newblock In \emph{Proceedings of the Annual Meeting of the Association for
  Computational Linguistics (ACL)}.

\bibitem[{Haghighi and Vanderwende(2009)}]{Haghighi:2009}
Aria Haghighi and Lucy Vanderwende. 2009.
\newblock \href {https://www.aclweb.org/anthology/N09-1041} {Exploring content
  models for multi-document summarization}.
\newblock In \emph{Proceedings of the North American Chapter of the Association
  for Computational Linguistics (NAACL)}.

\bibitem[{Hermann et~al.(2015)Hermann, Kocisky, Grefenstette, Espeholt, Kay,
  Suleyman, and Blunsom}]{Hermann:2015}
Karl~Moritz Hermann, Tomas Kocisky, Edward Grefenstette, Lasse Espeholt, Will
  Kay, Mustafa Suleyman, and Phil Blunsom. 2015.
\newblock \href {https://arxiv.org/abs/1506.03340} {Teaching machines to read
  and comprehend}.
\newblock In \emph{Proceedings of Neural Information Processing Systems
  (NIPS)}.

\bibitem[{Hinton et~al.(2018)Hinton, Sabour, and Frosst}]{Hinton:2018}
Geoffrey Hinton, Sara Sabour, and Nicholas Frosst. 2018.
\newblock \href {https://openreview.net/pdf?id=HJWLfGWRb} {Matrix capsules with
  {EM} routing}.
\newblock In \emph{Proceedings of the International Conference on Learning
  Representations (ICLR)}.

\bibitem[{Hirao et~al.(2013)Hirao, Yoshida, Nishino, Yasuda, and
  Nagata}]{Hirao:2013}
Tsutomu Hirao, Yasuhisa Yoshida, Masaaki Nishino, Norihito Yasuda, and Masaaki
  Nagata. 2013.
\newblock \href {https://www.aclweb.org/anthology/D13-1158} {Single-document
  summarization as a tree knapsack problem}.
\newblock In \emph{Proceedings of the Conference on Empirical Methods in
  Natural Language Processing (EMNLP)}.

\bibitem[{Hong et~al.(2014)Hong, Conroy, Favre, Kulesza, Lin, and
  Nenkova}]{Hong:2014}
Kai Hong, John~M Conroy, Benoit Favre, Alex Kulesza, Hui Lin, and Ani Nenkova.
  2014.
\newblock \href
  {http://www.lrec-conf.org/proceedings/lrec2014/pdf/1093_Paper.pdf} {A
  repository of state of the art and competitive baseline summaries for generic
  news summarization}.
\newblock In \emph{Proceedings of the Ninth International Conference on
  Language Resources and Evaluation (LREC)}.

\bibitem[{Hu et~al.(2019)Hu, Rudinger, Post, and Durme}]{Hu:2019}
J.~Edward Hu, Rachel Rudinger, Matt Post, and Benjamin~Van Durme. 2019.
\newblock \href {https://arxiv.org/abs/1901.03644} {{ParaBank}: {M}onolingual
  bitext generation and sentential paraphrasing via lexically-constrained
  neural machine translation}.
\newblock In \emph{Proceedings of the Association for the Advancement of
  Artificial Intelligence (AAAI)}.

\bibitem[{Kedzie et~al.(2018)Kedzie, McKeown, and III}]{Kedzie:2018}
Chris Kedzie, Kathleen McKeown, and Hal~Daume III. 2018.
\newblock \href {https://arxiv.org/abs/1810.12343} {Content selection in deep
  learning models of summarization}.
\newblock In \emph{Proceedings of the Conference on Empirical Methods in
  Natural Language Processing (EMNLP)}.

\bibitem[{Kim et~al.(2014)Kim, Sigal, and Xing}]{Kim:2014}
Gunhee Kim, Leonid Sigal, and Eric~P. Xing. 2014.
\newblock \href {https://ieeexplore.ieee.org/document/6909934/} {Joint
  summarization of large-scale collections of web images and videos for
  storyline reconstruction}.
\newblock In \emph{Proceedings of CVPR}.

\bibitem[{Knight and Marcu(2002)}]{Knight:2002}
Kevin Knight and Daniel Marcu. 2002.
\newblock \href {https://dl.acm.org/citation.cfm?id=604207} {Summarization
  beyond sentence extraction: {A} probabilistic approach to sentence
  compression}.
\newblock \emph{Artificial Intelligence}.

\bibitem[{Kryscinski et~al.(2018)Kryscinski, Paulus, Xiong, and
  Socher}]{Kryscinski:2018}
Wojciech Kryscinski, Romain Paulus, Caiming Xiong, and Richard Socher. 2018.
\newblock \href {https://arxiv.org/abs/1808.07913} {Improving abstraction in
  text summarization}.
\newblock In \emph{Proceedings of the Conference on Empirical Methods in
  Natural Language Processing (EMNLP)}.

\bibitem[{Kulesza and Taskar(2011)}]{Kulesza:2011}
Alex Kulesza and Ben Taskar. 2011.
\newblock \href {https://dl.acm.org/citation.cfm?id=3020597} {Learning
  determinantal point processes}.
\newblock In \emph{Proceedings of the Conference on Uncertainty in Artificial
  Intelligence (UAI)}.

\bibitem[{Kulesza and Taskar(2012)}]{Kulesza:2012}
Alex Kulesza and Ben Taskar. 2012.
\newblock \href {https://arxiv.org/abs/1207.6083} {\emph{Determinantal Point
  Processes for Machine Learning}}.
\newblock Now Publishers Inc.

\bibitem[{Lebanoff et~al.(2019)Lebanoff, Song, Dernoncourt, Kim, Kim, Chang,
  and Liu}]{Lebanoff:2019}
Logan Lebanoff, Kaiqiang Song, Franck Dernoncourt, Doo~Soon Kim, Seokhwan Kim,
  Walter Chang, and Fei Liu. 2019.
\newblock Scoring sentence singletons and pairs for abstractive summarization.
\newblock In \emph{Proceedings of the Annual Meeting of the Association for
  Computational Linguistics (ACL)}.

\bibitem[{Lebanoff et~al.(2018)Lebanoff, Song, and Liu}]{Lebanoff:2018}
Logan Lebanoff, Kaiqiang Song, and Fei Liu. 2018.
\newblock \href {https://aclweb.org/anthology/D18-1446} {Adapting the neural
  encoder-decoder framework from single to multi-document summarization}.
\newblock In \emph{Proceedings of the Conference on Empirical Methods in
  Natural Language Processing (EMNLP)}.

\bibitem[{Li et~al.(2013)Li, Liu, Weng, and Liu}]{Li:2013:EMNLP}
Chen Li, Fei Liu, Fuliang Weng, and Yang Liu. 2013.
\newblock \href {https://www.aclweb.org/anthology/D13-1047} {Document
  summarization via guided sentence compression}.
\newblock In \emph{Proceedings of the 2013 Conference on Empirical Methods in
  Natural Language Processing (EMNLP)}.

\bibitem[{Li et~al.(2014)Li, Liu, Liu, Zhao, and Weng}]{Li:2014:EMNLP}
Chen Li, Yang Liu, Fei Liu, Lin Zhao, and Fuliang Weng. 2014.
\newblock \href {https://www.aclweb.org/anthology/D14-1076} {Improving
  multi-document summarization by sentence compression based on expanded
  constituent parse tree}.
\newblock In \emph{Proceedings of the Conference on Empirical Methods on
  Natural Language Processing (EMNLP)}.

\bibitem[{Lin(2004)}]{Lin:2004}
Chin-Yew Lin. 2004.
\newblock \href {https://www.aclweb.org/anthology/W04-1013} {{ROUGE}: a package
  for automatic evaluation of summaries}.
\newblock In \emph{Proceedings of ACL Workshop on Text Summarization Branches
  Out}.

\bibitem[{Lin and Bilmes(2010)}]{Lin:2010:NAACL}
Hui Lin and Jeff Bilmes. 2010.
\newblock \href {https://aclweb.org/anthology/N10-1134} {Multi-document
  summarization via budgeted maximization of submodular functions}.
\newblock In \emph{Proceedings of NAACL}.

\bibitem[{Luo and Litman(2015)}]{Luo:2015}
Wencan Luo and Diane Litman. 2015.
\newblock \href {https://www.aclweb.org/anthology/D15-1227} {Summarizing
  student responses to reflection prompts}.
\newblock In \emph{Proceedings of the Conference on Empirical Methods in
  Natural Language Processing (EMNLP)}.

\bibitem[{Luo et~al.(2016)Luo, Liu, Liu, and Litman}]{Luo:2016:NAACL}
Wencan Luo, Fei Liu, Zitao Liu, and Diane Litman. 2016.
\newblock \href {https://www.aclweb.org/anthology/N16-1010} {Automatic
  summarization of student course feedback}.
\newblock In \emph{Proceedings of the North American Chapter of the Association
  for Computational Linguistics: Human Language Technologies (NAACL)}.

\bibitem[{Martins and Smith(2009)}]{Martins:2009}
Andre F.~T. Martins and Noah~A. Smith. 2009.
\newblock \href {https://www.aclweb.org/anthology/W09-1801} {Summarization with
  a joint model for sentence extraction and compression}.
\newblock In \emph{Proceedings of the ACL Workshop on Integer Linear
  Programming for Natural Language Processing}.

\bibitem[{Nallapati et~al.(2017)Nallapati, Zhai, and Zhou}]{Nallapati:2017}
Ramesh Nallapati, Feifei Zhai, and Bowen Zhou. 2017.
\newblock \href {https://arxiv.org/abs/1611.04230} {{SummaRuNNer}: {A}
  recurrent neural network based sequence model for extractive summarization of
  documents}.
\newblock In \emph{Proceedings of the Thirty-First AAAI Conference on
  Artificial Intelligence (AAAI)}.

\bibitem[{Nallapati et~al.(2016)Nallapati, Zhou, dos Santos, Gulcehre, and
  Xiang}]{Nallapati:2016}
Ramesh Nallapati, Bowen Zhou, Cicero dos Santos, Caglar Gulcehre, and Bing
  Xiang. 2016.
\newblock \href {https://arxiv.org/abs/1602.06023} {Abstractive text
  summarization using sequence-to-sequence rnns and beyond}.
\newblock In \emph{Proceedings of SIGNLL}.

\bibitem[{Narayan et~al.(2018)Narayan, Cohen, and Lapata}]{Narayan:2018}
Shashi Narayan, Shay~B. Cohen, and Mirella Lapata. 2018.
\newblock \href {https://arxiv.org/abs/1802.08636} {Ranking sentences for
  extractive summarization with reinforcement learning}.
\newblock In \emph{Proceedings of the 16th Annual Conference of the North
  American Chapter of the Association for Computational Linguistics: Human
  Language Technologies (NAACL-HLT)}.

\bibitem[{Nenkova and McKeown(2011)}]{Nenkova:2011}
Ani Nenkova and Kathleen McKeown. 2011.
\newblock \href {https://www.nowpublishers.com/article/Details/INR-015}
  {Automatic summarization}.
\newblock \emph{Foundations and Trends in Information Retrieval}.

\bibitem[{Over and Yen(2004)}]{Over:2004}
Paul Over and James Yen. 2004.
\newblock \href {https://duc.nist.gov/pubs/2004slides/duc2004.intro.pdf} {An
  introduction to {DUC}-2004}.
\newblock \emph{National Institute of Standards and Technology}.

\bibitem[{Paulus et~al.(2017)Paulus, Xiong, and Socher}]{Paulus:2017}
Romain Paulus, Caiming Xiong, and Richard Socher. 2017.
\newblock \href {https://arxiv.org/abs/1705.04304} {A deep reinforced model for
  abstractive summarization}.
\newblock In \emph{Proceedings of the Conference on Empirical Methods in
  Natural Language Processing (EMNLP)}.

\bibitem[{Rush et~al.(2015)Rush, Chopra, and Weston}]{Rush:2015}
Alexander~M. Rush, Sumit Chopra, and Jason Weston. 2015.
\newblock \href {https://www.aclweb.org/anthology/D15-1044} {A neural attention
  model for sentence summarization}.
\newblock In \emph{Proceedings of EMNLP}.

\bibitem[{Sabour et~al.(2017)Sabour, Frosst, and Hinton}]{Sabour:2017}
Sara Sabour, Nicholas Frosst, and Geoffrey~E. Hinton. 2017.
\newblock \href
  {https://papers.nips.cc/paper/6975-dynamic-routing-between-capsules.pdf}
  {Dynamic routing between capsules}.
\newblock In \emph{Proceedings of the 31st Conference on Neural Information
  Processing Systems (NIPS)}.

\bibitem[{See et~al.(2017)See, Liu, and Manning}]{See:2017}
Abigail See, Peter~J. Liu, and Christopher~D. Manning. 2017.
\newblock \href {https://arxiv.org/abs/1704.04368} {Get to the point:
  {S}ummarization with pointer-generator networks}.
\newblock In \emph{Proceedings of the Annual Meeting of the Association for
  Computational Linguistics (ACL)}.

\bibitem[{Shen and Li(2010)}]{Shen:2010}
Chao Shen and Tao Li. 2010.
\newblock \href {https://www.aclweb.org/anthology/C10-1111} {Multi-document
  summarization via the minimum dominating set}.
\newblock In \emph{Proceedings of the International Conference on Computational
  Linguistics (COLING)}.

\bibitem[{Song et~al.(2018)Song, Zhao, and Liu}]{Song:2018}
Kaiqiang Song, Lin Zhao, and Fei Liu. 2018.
\newblock \href {https://aclweb.org/anthology/C18-1146} {Structure-infused copy
  mechanisms for abstractive summarization}.
\newblock In \emph{Proceedings of the International Conference on Computational
  Linguistics (COLING)}.

\bibitem[{Takamura and Okumura(2009)}]{Takamura:2009}
Hiroya Takamura and Manabu Okumura. 2009.
\newblock \href {https://www.aclweb.org/anthology/E09-1089} {Text summarization
  model based on maximum coverage problem and its variant}.
\newblock In \emph{Proceedings of the European Chapter of the Association for
  Computational Linguistics (EACL)}.

\bibitem[{Tan et~al.(2017)Tan, Wan, and Xiao}]{Tan:2017}
Jiwei Tan, Xiaojun Wan, and Jianguo Xiao. 2017.
\newblock \href {https://www.aclweb.org/anthology/P17-1108} {Abstractive
  document summarization with a graph-based attentional neural model}.
\newblock In \emph{Proceedings of the Annual Meeting of the Association for
  Computational Linguistics (ACL)}.

\bibitem[{Tarnpradab et~al.(2017)Tarnpradab, Liu, and Hua}]{Tarnpradab:2017}
Sansiri Tarnpradab, Fei Liu, and Kien~A. Hua. 2017.
\newblock \href {https://arxiv.org/abs/1805.10390} {Toward extractive
  summarization of online forum discussions via hierarchical attention
  networks}.
\newblock In \emph{Proceedings of the 30th Florida Artificial Intelligence
  Research Society Conference (FLAIRS)}.

\bibitem[{Thadani and McKeown(2013)}]{Thadani:2013}
Kapil Thadani and Kathleen McKeown. 2013.
\newblock \href {https://www.aclweb.org/anthology/W13-3508} {Sentence
  compression with joint structural inference}.
\newblock In \emph{Proceedings of CoNLL}.

\bibitem[{Vanderwende et~al.(2007)Vanderwende, Suzuki, Brockett, and
  Nenkova}]{Vanderwende:2007}
Lucy Vanderwende, Hisami Suzuki, Chris Brockett, and Ani Nenkova. 2007.
\newblock \href {https://www.cis.upenn.edu/~nenkova/papers/ipm.pdf} {Beyond
  {SumBasic}: {T}ask-focused summarization with sentence simplification and
  lexical expansion}.
\newblock \emph{Information Processing and Management}, 43(6):1606--1618.

\bibitem[{Wang et~al.(2013)Wang, Raghavan, Castelli, Florian, and
  Cardie}]{Wang:2013}
Lu~Wang, Hema Raghavan, Vittorio Castelli, Radu Florian, and Claire Cardie.
  2013.
\newblock \href {https://arxiv.org/abs/1606.07548} {A sentence compression
  based framework to query-focused multi-document summarization}.
\newblock In \emph{Proceedings of ACL}.

\bibitem[{Williams et~al.(2018)Williams, Nangia, and Bowman}]{Williams:2018}
Adina Williams, Nikita Nangia, and Samuel~R. Bowman. 2018.
\newblock \href {https://arxiv.org/abs/1704.05426} {A broad-coverage challenge
  corpus for sentence understanding through inference}.
\newblock In \emph{Proceedings of the North American Chapter of the Association
  for Computational Linguistics (NAACL)}.

\bibitem[{Yang et~al.(2018)Yang, Qu, Shen, Liu, Zhao, and Zhu}]{Yang:2018}
Min Yang, Qiang Qu, Ying Shen, Qiao Liu, Wei Zhao, and Jia Zhu. 2018.
\newblock \href {https://www.aclweb.org/anthology/C18-1095} {Aspect and
  sentiment aware abstractive review summarization}.
\newblock \emph{Proceedings of the International Conference on Computational
  Linguistics (COLING)}.

\bibitem[{Yasunaga et~al.(2017)Yasunaga, Zhang, Meelu, Pareek, Srinivasan, and
  Radev}]{Yasunaga:2017}
Michihiro Yasunaga, Rui Zhang, Kshitijh Meelu, Ayush Pareek, Krishnan
  Srinivasan, and Dragomir Radev. 2017.
\newblock Graph-based neural multi-document summarization.
\newblock In \emph{Proceedings of the Conference on Computational Natural
  Language Learning (CoNLL)}.

\bibitem[{Yogatama et~al.(2015)Yogatama, Liu, and Smith}]{Yogatama:2015:EMNLP}
Dani Yogatama, Fei Liu, and Noah~A. Smith. 2015.
\newblock \href {https://www.aclweb.org/anthology/D15-1228} {Extractive
  summarization by maximizing semantic volume}.
\newblock In \emph{Proceedings of the Conference on Empirical Methods on
  Natural Language Processing (EMNLP)}.

\bibitem[{Zajic et~al.(2007)Zajic, Dorr, Lin, and Schwartz}]{Zajic:2007}
David Zajic, Bonnie~J. Dorr, Jimmy Lin, and Richard Schwartz. 2007.
\newblock \href
  {http://users.umiacs.umd.edu/~jimmylin/publications/Zajic_etal_IPM2007.pdf}
  {Multi-candidate reduction: {S}entence compression as a tool for document
  summarization tasks}.
\newblock \emph{Information Processing and Management}.

\bibitem[{Zhang et~al.(2016)Zhang, Chao, Sha, and Grauman}]{Zhang:2016:DPP}
Ke~Zhang, Wei-Lun Chao, Fei Sha, and Kristen Grauman. 2016.
\newblock \href {https://arxiv.org/abs/1605.08110} {Video summarization with
  long short-term memory}.
\newblock In \emph{Proceedings of the European Conference on Computer Vision
  (ECCV)}.

\bibitem[{Zhao et~al.(2019)Zhao, Peng, Eger, Cambria, and
  Yang}]{Zhao:2019:Capsnet}
Wei Zhao, Haiyun Peng, Steffen Eger, Erik Cambria, and Min Yang. 2019.
\newblock Towards scalable and generalizable capsule network and its {NLP}
  applications.
\newblock In \emph{Proceedings of the Annual Meeting of the Association for
  Computational Linguistics (ACL)}.

\bibitem[{Zhao et~al.(2018)Zhao, Ye, Yang, Lei, Zhang, and
  Zhao}]{Zhao:2018:Capsnet}
Wei Zhao, Jianbo Ye, Min Yang, Zeyang Lei, Soufei Zhang, and Zhou Zhao. 2018.
\newblock \href {https://arxiv.org/abs/1804.00538} {Investigating capsule
  networks with dynamic routing for text classification}.
\newblock In \emph{Proceedings of the Conference on Empirical Methods in
  Natural Language Processing (EMNLP)}.

\bibitem[{Zhou et~al.(2017)Zhou, Yang, Wei, and Zhou}]{Zhou:2017}
Qingyu Zhou, Nan Yang, Furu Wei, and Ming Zhou. 2017.
\newblock \href {https://arxiv.org/abs/1704.07073} {Selective encoding for
  abstractive sentence summarization}.
\newblock In \emph{Proceedings of the Annual Meeting of the Association for
  Computational Linguistics (ACL)}.

\end{thebibliography}
\bibliographystyle{acl_natbib}

\newpage
\appendix

\section{Supplemental Material}
\label{sec:supplemental}

In this section we summarize the hyperparameters used for the capsule networks.
They include:
the embedding size $\textsf{E}$ is set to 300 dimensions;
the maximum sentence length $\textsf{L}$ is 44 words;
in the convolutional layer we use $d$=100 filters for each filter size, and there are 5 filter sizes in total: $k \in \{3, 4, 5, 6, 7\}$.
The number of high-level capsules $\textsf{M}$ is set to 12, and the dimension of capsules $\textsf{B}$ is set to 30, both are tuned on the development set.
The dynamic routing process is repeated for $r$=3 iterations, following~\cite{Sabour:2017}.
Further, the coefficient $\lambda$ for the reconstruction loss term is set to 5e-5.
We use a vocabulary of 50K words for reconstructing the sentences; they are the most frequently appearing words of the dataset.

\end{document}